\documentclass[letterpaper]{article} 
\usepackage{aaai24}  
\usepackage{times}  
\usepackage{helvet}  
\usepackage{courier}  
\usepackage[hyphens]{url}  
\usepackage{graphicx} 
\usepackage{natbib}  
\usepackage{caption} 
\usepackage{algorithm}
\usepackage[noend]{algpseudocode}

\usepackage{amsmath, amsthm, amssymb}
\usepackage{booktabs} 
\usepackage{multirow}
\usepackage{caption}
\usepackage{lipsum}
\usepackage{subcaption}

\theoremstyle{plain}

\theoremstyle{definition}
\newtheorem{definition}{Definition}

\usepackage{xcolor}
\newcommand{\fref}[1]{Fig.~\ref{#1}}
\newcommand{\tref}[1]{Table~\ref{#1}}

\usepackage{newfloat}
\usepackage{listings}
\DeclareCaptionStyle{ruled}{labelfont=normalfont,labelsep=colon,strut=off} 
\floatstyle{ruled}
\newfloat{listing}{tb}{lst}{}
\floatname{listing}{Listing}
\title{LRS: Enhancing Adversarial Transferability through Lipschitz Regularized Surrogate}
\author{
    Tao Wu\textsuperscript{\rm 1},
    Tie Luo\textsuperscript{\rm 1}\thanks{Corresponding author.}, 
    Donald C. Wunsch II\textsuperscript{\rm 2}
}
\affiliations{
    \textsuperscript{\rm 1}Department of Computer Science, Missouri University of Science and Technology\\
    \textsuperscript{\rm 2}Department of Electrical and Computer Engineering, Missouri University of Science and Technology\\

    \{wuta, tluo, dwunsch\}@mst.edu
}

\usepackage{bibentry}

\begin{document}

\maketitle

\begin{abstract}
The transferability of adversarial examples is of central importance to transfer-based black-box adversarial attacks. Previous works for generating transferable adversarial examples focus on attacking \emph{given} pretrained surrogate models while the connections between surrogate models and adversarial trasferability have been overlooked. In this paper, we propose {\em Lipschitz Regularized Surrogate} (LRS) for transfer-based black-box attacks, a novel approach that transforms surrogate models towards favorable adversarial transferability. Using such transformed surrogate models, any existing transfer-based black-box attack can run without any change, yet achieving much better performance. Specifically, we impose Lipschitz regularization on the loss landscape of surrogate models to enable a smoother and more controlled optimization process for generating more transferable adversarial examples. In addition, this paper also sheds light on the connection between the inner properties of surrogate models and adversarial transferability, where three factors are identified: smaller local Lipschitz constant, smoother loss landscape, and stronger adversarial robustness. We evaluate our proposed LRS approach by attacking state-of-the-art standard deep neural networks and defense models. The results demonstrate significant improvement on the attack success rates and transferability. Our code is available at https://github.com/TrustAIoT/LRS.
\end{abstract}

\section{Introduction}

Deep Neural Networks (DNNs) are the workhorse of a broad variety of computer vision tasks and have made resounding success in classification \cite{he2016resnet}, object detection \cite{redmon2016you}, segmentation \cite{ronneberger2015u}, and so on. However, they are vulnerable to \emph{adversarial examples} (AE), which are data samples that are perturbed by human-imperceptible noises yet result in misclassifications. This lack of adversarial robustness can cause serious safety and security consequences in applications such as healthcare, neuroscience, finance, self-driving, and reconnaissance, to name a few.

Adversarial attacks are commonly launched under two settings, white-box and black-box attacks. In the white-box setting, adversaries have full knowledge of target models, including model structures, parameters and weights, data and loss functions used to train the models. Therefore, they can add such perturbation to benign images that the loss on the perturbed images is maximized. An efficient way to do this involves iteratively incorporating the gradient of the loss w.r.t. input \cite{goodfellow2015FGSM,madry2018pgd} into perturbations. White-box attacks are important for evaluating and developing robust models, and also serve as the backend method for many black-box attacks. However, they are limited by the requirement of having to know the internal details of target models. In the black-box setting, adversaries do not need insider knowledge about target models other than their external interface (type/format of input and output), and usually take two types of approaches, query-based or transfer-based. Query-based approaches attempt to estimate the gradients of a target model's loss function by querying it with a large number of input samples and inspecting the outputs. Such frequent queries make it easy to be detected and defend them. On the other hand, transfer-based approaches use {\em surrogate} models to generate {\em transferable} AE which can attack a wide range of models, and hence are more effective to form stronger and more covert black-box attacks.

\begin{figure}[t]
    \centering
    \includegraphics[width=\linewidth]{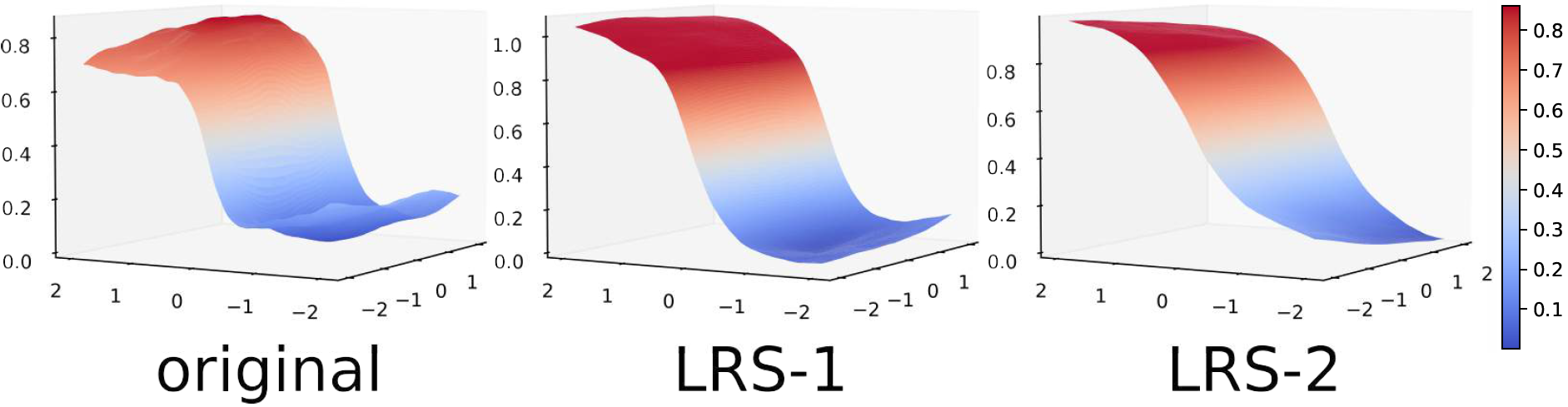}
    \caption{The loss landscape of original and transformed surrogate model: corrugated vs. smooth. Transformed surrogate models offer more stable input gradients and make the generated AE more generalizable, enabling more potent attacks.}
    \label{fig:loss_landscape}
\end{figure}

The \emph{transferability} of AE is of central importance for transfer-based attacks. Unveiling principles of adversarial transferability provides insight into understanding the working mechanism of DNNs and designing robust DNNs. In the literature, several directions have been explored to improve the transferability of AE from the attackers' perspective. These include optimization-based \cite{dong2018boosting,lin2020nesterov}, smoothing-based \cite{dong2019evading,xie2019improving,wu2019skip,guo2020backpropagating}, attention-based \cite{wu2020boosting}, and ensemble-based \cite{li2020learning} methods. Despite these efforts, a large gap of attack success rate still exists between the transfer-based black-box setting and the ideal white-box setting. The major reason is that AE created on a surrogate model can easily be trapped into the surrogate model's exclusive blind spots, resulting in poor generalization to fool other target models --- a phenomenon known as AE \emph{overfitting}.

Prior work on boosting adversarial transferability has focused on the AE crafting process itself, either by (1) manipulating the input images \cite{xie2019improving} or their attention maps \cite{inkawhich2019feature}, or (2) tuning the AE optimization steps such as applying momentum \cite{dong2018boosting} or variance reduction \cite{xiong2022stochastic}. However, the surrogate model, on which AE crafting is hinged, has been taken as given and not adequately explored. Specifically, what internal properties of a surrogate model are important to produce transferable AE, and (how) are they achievable? Answering this question points toward a new direction to adversarial machine learning.

We were inspired by the intricate terrain of the loss landscape w.r.t. inputs, which is characterized by peaks, valleys, and plateaus, profoundly influencing the behavior of optimization algorithms that generate AE. Thus, we propose to impose local Lipschitz regularization on the loss landscape of {\em surrogate models}, striving to alleviate notorious challenges in optimization posed by sharp gradients, vanishing or exploding gradients, and chaotic oscillations of gradient descent within the loss landscape. 
Upon that, the optimization process can traverse terrains with ease, not encountering steep slopes, cliffs, narrow valleys, etc., thereby allowing for creating stronger (i.e., more generalizable) transfer-based black-box attacks. As shown in \fref{fig:loss_landscape}, such regularized surrogate models offer more stable input gradients and flatter local optima which help avoid AE overfitting and create more transferable AE.

The contributions of this paper are summarized below:

\begin{itemize}

    \item Unlike prior work which all focuses on the AE generation process per se, we transform surrogate models on which that process is based, such that any existing transfer-based black-box AE generation methods can simply run on our LRS-transformed surrogate models, like a ``cushion'', without any change yet achieving much better performance.

    \item To the best of our knowledge, this is the first work that establishes a connection between the inner properties of surrogate models and AE transferability. We identify three such properties that would favor adversarial transferability, namely smaller local Lipschitz constant, smoother loss landscape, and stronger adversarial robustness, offering further insights into understanding adversarial transferability.
    
    \item We conduct extensive evaluation on ImageNet and demonstrate that, by applying LRS to a basic AE generation method (PGD), it yields superior adversarial transferability for 7 state-of-the-art black-box attacks on 10 target models.
\end{itemize}

\section{Related Work}

\subsection {Adversarial Transferability}
The transferability of adversarial examples enables transfer-based black-box attacks \cite{szegedy2014intriguing}. Such attacks require the least knowledge of target models and thus often pose the biggest threat to AI systems deployed in the real world. This black-box approach is to apply white-box attacks on surrogate models to find adversarial examples that can fool as many black-box target models as possible, known as transferability of the AE. Many works have been proposed to improve the transferability of AE. \textbf{Optimization-based approaches} focus on finding direction of the gradients towards optima that lead to better transferability. For example, Momentum Iterative Method (MIM) \cite{dong2018boosting} integrates a momentum term into the gradient calculation to stabilize the update direction. Reverse Adversarial Perturbation (RAP) \cite{qin2022boosting} seeks targeted AE located at a region with uniformly low loss value. \textbf{Smoothing-based approaches} smooth gradients by averaging gradients from multiple datapoints around the current AE. Diverse Inputs Method (DIM) \cite{xie2019improving} averages the gradients of randomly resized and padded inputs to generate AE. Translation-invariant Attack (TIM) \cite{dong2019evading}, Scale Invariance Attack (SIM) \cite{lin2020nesterov}, Smoothed Gradient Attack (SGM) \cite{wu2020towards}, and Admix Attack (Admix) \cite{wang2021admix} also fall into this category. \textbf{Attention-based approaches} modify the important features in attention maps, motivated by the observation that different deep networks classify the same image based on similar important features. For instance, Attention guided Transfer Attack (ATA) \cite{wu2020boosting} uses the gradients of an objective function w.r.t. neuron outputs to derive an attention map and seek AE that maximizes the difference between its attention map and the corresponding benign sample's map. Similar approaches include Jacobian based Saliency Map Attack (JSMA) \cite{papernot2016limitations}, Attack on Attention (AoA) \cite{chen2020universal} and Activation attack (AA) \cite{inkawhich2019feature}. \textbf{Ensemble-based approaches} take advantage of an ensemble of surrogate models with the belief that if an AE can attack multiple models, then it is more likely to transfer to other models as well. For instance, \cite{liu2017delving} proposes to generate AE on an ensemble of models with different architectures. Large Geometric Vicinity (LGV) \cite{gubri2022lgv} collects multiple checkpoints along the training trajectory, on which the  attack was performed on an ensemble of these models. \cite{li2023making} develops an ensemble attack from a Bayesian formulation which samples multiple models from the posterior distribution of parameter space.
 
\subsection{Connection with Surrogate Model's Geometry}
Most previous works for boosting adversarial transferability are based on fixed pre-trained surrogate models, while little attention has been paid to exploring what properties of the surrogate models would enable more adversarial transferability, and whether/how to change them. One related work is \cite{wu2018understanding}, which studied some model aspects such as network architecture and model capacity with {\em given} pretrained models. Recently, \cite{charles2019geometric,tramer2017space} theoretically analyzed the neural network geometry in relation to adversarial transferability but no particular method was designed or proposed and no Lipchitz properties were investigated.

In this paper, we contend that imposing Lipschitz regularization on surrogate models and hence changing this ``foundation slate'' on which black-box AE are crafted can enhance adversarial transferability. Regularization toward smoothness has been particularly successful in the design of GANs \cite{gulrajani2017improved}, but the connection between local Lipschitzness and adversarial transferability has never been explored. Furthermore, we convert this conceptual idea into a concrete method with rigorous theoretical characterizations and empirical evaluations.

\section{Methodology} \label{sec:methodology}

Given a classification model $f(x): x \in \mathcal{X} \rightarrow y \in \mathcal{Y}$ that outputs a predicted label $y$ for an input $x$, we aim to craft an adversarial example $x^*$ which is visually indistinguishable from $x$ but will be misclassified by the classifier, i.e., $f\left(x^*\right) \neq y$. This objective can be formulated as the following optimization problem:
\vspace{-1mm}
\begin{align} \label{eq:1}
    \arg \max _{x^*} \ell\left(x^*, y\right), \;\; \text {s.t. }\left\|x^*-x\right\|_p \leq \epsilon,
\end{align}
where the loss function $\ell(\cdot, \cdot)$ is often the cross-entropy loss, and the $l_p$-norm measures the discrepancy between $x$ and $x^*$. We adopt $p=\infty$ as is common in the literature. Optimizing Eq.~\eqref{eq:1} needs to calculate the gradient of the loss function, which unfortunately is not accessible in the black-box setting. Therefore, we seek a surrogate model on which we aim to create transferable AE that can attack many other unknown target models.

The choice of surrogate model plays a critical role in generating transferable AE. However, previous works have focused on {\em selecting} pretrained surrogate models in terms of network architecture, model capacity and accuracy \cite{wu2020towards}, and 
attacking them \emph{as given}. Those models' internal properties such as loss geometry and robustness have been overlooked. In our work, we set to alter any given surrogate model towards desired internal properties that favor adversarial transferability.

\subsection{LRS-1: Lipschitz Regularization on the First Order of Loss Landscape}

\begin{definition} \label{def:lip_continous}
A function $f(x)$ is locally \emph{$L_c$-Lipschitz continuous} on an open set $\Omega \subset \mathbb{R}^m$  if there exists a constant $0 \leq L_c<\infty$ satisfying
\begin{align*}
  \forall x_1, x_2 \in \Omega,\ \|f(x_1) - f(x_2)\|_2 \leq L_c\,\|x_1 - x_2\|_2.
\end{align*}

\end{definition}

The smallest $L_c$ for which the above inequality is satisfied is called the {\em Lipschitz constant} of $f(\cdot)$. Without loss of generality, we assume that the loss function of surrogate model is a locally Lipschitz function around a datapoint $x$ (i.e., in the neighborhood $\mathcal{B}_\epsilon(x)=\{ x^{\prime}:\left\|x-x^{\prime}\right\|_2 \leq \epsilon \}$). Our aim is to restrict the local Lipschitz constant $L_c$. The rationale is that if the loss function of the surrogate model has a small local Lipschitz constant $L_c$, the change of loss will be small in the neighborhood of $x$; thus for any adversarial examples $x^*$ that incurs a large loss $\ell(x^*)$, datapoints around $x^*$ are also likely to incur large loss, and hence tend to be adversarial on {\em other} unknown target models as well since neural network classifiers generally share similar decision boundaries and loss landscape \cite{liu2017delving}. 

To constrain the local Lipschitz constant $L_c$, we derive a regularization term that can reshape the loss landscape of surrogate models towards the above goal. According to the mean value theorem, for all $x^{\prime} \in \mathcal{B}_\epsilon(x)$,
\begin{align}
\left\|\ell\left({x}^{\prime}\right)-\ell\left({x}\right)\right\|_2 = \left\|\nabla \ell({\zeta})\left({x}^{\prime}-{x}\right)\right\|_2,
\end{align}
where ${\zeta}=c {x}+(1-c) {x}^{\prime}, c \in[0,1]$. Then the Cauchy-Schwarz inequality gives that
\begin{align}
\left\|\ell\left({x}^{\prime}\right)-\ell\left({x}\right)\right\|_2 \leq\|\nabla \ell({\zeta})\|_2\left\|\left({x}^{\prime}-{x}\right)\right\|_2.
\end{align}

When $x^{\prime} \rightarrow x$, the corresponding Lipschitz constant $L_c=\left\|\nabla \ell({\zeta})\right\|_2$ approximates to $\left\|\nabla \ell\left({x}\right)\right\|_2$. Therefore, we transform our original aim of constraining the Lipschitz constant $L_c$ into constraining $\left\|\nabla \ell\left({x}\right)\right\|_2$ so that the crafted AE would reach a smoother and flatter optimum when maximizing the loss.

To this end, we impose the constraint of small Lipscitz constant to the loss of surrogate model by optimizing the following new objective:
\begin{align}\label{eq:L1}
L({x, y}) = \ell(x, y) + \lambda_1\left\|\nabla_x \ell(x, y)\right\|_{2}^{2}
\end{align}
where $\ell(\cdot)$ is the original loss function of the surrogate model, and we square the gradient norm in order to penalize more on larger norms.

\begin{algorithm}[t]
\caption{LRS-1 (using PGD as an example base)}
\label{alg}
\begin{flushleft}
    \textbf{Input:} A clean sample $x$ with ground-truth label $y$; a pretrained surrogate model $f(\cdot)$;\\
    \textbf{Hyper-parameters:} Finetune epochs $n$; batch size $m$; learning rate $\eta$; training dataset $D$; step size $h$; perturbation size $\epsilon$; maximum iterations $T$;  regularization coefficient $\lambda$\\
    \textbf{Output:} A transferable AE $x^{adv}$
\end{flushleft}
\begin{algorithmic}[1]
    \State Pretrained surrogate model $f_0$ with weight $w_0$
    \For{epoch $=0$ to $n-1$}
    \For{t $=0$ to $len(D)/m$}
        \State sample minibatch $\left\{\left(x_{i}, y_{i}\right)\right\}_{i=1, \ldots, m}$
        \State $g_{i}=\nabla_x \ell\left(x_{i}, y_{i} ; w_t\right)$
        \State $d_{i}=\operatorname{sign}(g_{i})$
        \State $z_{i}=x_{i}+h d_{i}$
        \State $\mathcal{L}(w_t)=\sum_{i=1}^m \ell\left(x_i, y_i ; w_t\right)$
        \State $\mathcal{R}(w_t)=\sum_{i=1}^m \left(\ell\left(z_i, y_i ; w_t\right)-\ell\left(x_i, y_i ; w_t\right)\right)^2$
        \State $w_{t+1} = w_t-\frac{1}{m} \eta \nabla_w\left(\mathcal{L}\left(w_t\right)+\frac{1}{h^2} \lambda \mathcal{R}\left(w_t\right)\right)$
    \EndFor
    \EndFor
    \State \textbf{save} finetuned surrogate model $f_n$ with weight $w_n$
    \vspace{-8mm}
    \[\]
    \State $\alpha=\epsilon/T$; $x_{0}^{adv}=x$
    \For{$t=0$ to $T-1$}
        \State $g_t=\nabla_{x} \ell(x, w_n)$
        \State $x_{t+1}^{adv} = x_t^{adv} + \alpha \cdot \text{sign}(g_{t})$
        \State $x_{t+1}^{adv} = \operatorname{clip}\left(x_{t+1}^{adv}, 0,1\right)$
    \EndFor
    \State \textbf{return} $x^{adv}=x_{T}^{adv}$
\end{algorithmic}

\end{algorithm}

\begin{table*}[ht!]
\begin{center}
\resizebox{0.9\linewidth}{!}{
\begin{tabular}{|c|c|c|cccccc|}
\hline
$\epsilon$ & Transformed? & DenseNet* & VGG19 & ResNet18 & WRN & ResNeXt   & PyramidNet  & Average  \\
\hline\hline
 \multirow{4}{*}{4/255} & No & 100.00\% & 29.79\% & 19.04\% & 54.41\% & 69.41\% & 21.53\% & 38.84\% \\
 & LRS-1  & 99.73\% & 55.97\% & 42.16\% & 72.66\% & 80.93\% & 42.64\% & 58.87\%\\
 & LRS-2  & 99.82\% & 59.86\% & 48.98\% & 77.81\%  & 88.63\% & 46.78\% &  64.21\%\\
 & LRS-F  & 99.93\% & \textbf{65.16\%} & \textbf{54.23\%} & \textbf{81.49\%}  & \textbf{92.76\%} & \textbf{51.07\%} & \textbf{68.94\%}\\
\hline
 \multirow{4}{*}{8/255} & No & 100.00\% & 60.13\% & 35.54\% & 86.41\% & 95.60\% & 46.62\% & 64.85\% \\
 & LRS-1  & 100.00\% & 93.41\% & 77.82\% & 98.79\% & 99.75\% & 88.09\% &91.57\%\\
 & LRS-2  & 100.00\% & 95.26\% & 81.43\% & 99.27\%  & \textbf{99.87} \% & 92.69\% & 93.71\%\\
 & LRS-F  & 100.00\% & \textbf{96.21\%} & \textbf{86.41\%} & \textbf{99.45\%} & 99.84\% & \textbf{95.46\%} & \textbf{95.48}\%\\
\hline
\end{tabular}
}
\caption{Attack success rates of adversarial examples crafted on CIFAR10 dataset using original and transformed surrogate model under $\ell_\infty$ constraint with $\epsilon=4/255$ and $\epsilon=8/255$, PGD serves as the backbone method. `*' denotes white-box attacks.}
\label{tab:quick_exp}
\end{center}
\end{table*}

\subsection{LRS-2: Lipschitz Regularization on the Second Order of Loss Landscape}

\begin{definition} \label{def:lip_smooth}
A function $f(x)$ is said to have a Lipschitz continuous gradient on an open set $\Omega \subset \mathbb{R}^m$  if there exists a constant $0 \leq L_s<\infty$ satisfying
\begin{align*}
  \forall x_1, x_2 \in \Omega,\ \|\nabla f(x_1) - \nabla f(x_2)\|_2 \leq L_s\,\|x_1 - x_2\|_2.
\end{align*}

\end{definition}

From the convex optimization theory, we know that for a twice differentiable strongly convex $f(\cdot)$, the largest eigenvalue of the Hessian of $f$ is uniformly upper bounded by $L_s$ everywhere on $\Omega$. That is,
\begin{align}
L_s I \succeq \nabla^2 f(x)
\end{align}

Our aim is to restrict the Lipschitz continuous gradient of $f$, such that the largest eigenvalue of the Hessian of $f$ will be small. The rationale is that the local curvature geometry of a function is measured by its Hessian, whose eigenvectors and eigenvalues describe the directions of principal curvature and the amount of curvature in each direction, respectively. Thus, limiting the eigenvalues will lead to smaller curvature which translates to a more linear behaviour of the surrogate network. Besides, this regularization penalizes a steep loss surface, encouraging the optimization to move towards regions of flatter curvature, where the generated AE will have a better ability to generalize to new, unseen models \cite{qin2022boosting}. 

Thus, we propose a regularization on the second-order of the loss landscape as follows by ``linearlizing'' the surrogate model:
\begin{align}\label{eq:L2}
L({x, y}) = \ell(x, y) + \lambda_2\left\|\nabla_{x}^{2} \ell(x, y)\right\|_{2}^{2}.
\end{align}

\underline{\bf Remark:}
Note that the above two regularization formulations \eqref{eq:L1} \eqref{eq:L2} concern local Lipschitzness with respect to the {\em input space} instead of the {\em parameter space}. This is an important distinction from conventional neural network optimization.

\subsection{Optimizing the Regularized Loss}

In view of practical implementation, we also consider reducing the computational overhead and make our attack scalable to large neural networks and datasets. To this end, instead of training a surrogate model using our proposed regularized objective from scratch, we propose to fine-tune a pretrained network with only a few extra epochs (10 epochs in our implementation).

To efficiently calculate the regularization terms in \eqref{eq:L1} and \eqref{eq:L2}, we approximate them with finite difference methods (FDM), because computing the full Hessian matrix would incur prohibitive cost for high-dimensional datasets.

Let $d$ be the input gradient direction, i.e., $d=\operatorname{sign}(\nabla_x \ell(x, y))$, 
$h$ be the finite difference step size. Then, the input gradient norm is approximated by
\begin{align}
\left\|\nabla_x \ell(x, y)\right\|_2^2 \approx\left(\frac{\ell(x+h_1 d, y)-\ell(x, y)}{h_1}\right)^2
\end{align}

Similary, 
\begin{align}
\left\|\nabla_{x}^{2} \ell(x, y)\right\|_2^2 \approx\left(\frac{\nabla_{x}\ell(x+h_2 d, y)-\nabla_{x}\ell(x, y)}{h_2}\right)^2
\end{align}

This approximation significantly reduces the overhead of computing the Hessian directly. Moreover, it also allows us to harness an additional benefit of controlling large variations of loss and gradient, via the step size $h$ which specifies the neighborhood size of datapoint $x$.

Algorithm \ref{alg} presents LRS-1 in its entirety, employing Projected Gradient Descent (PGD) \cite{madry2018pgd} as a simple base to substantiate the attack. Notably, LRS serves as a versatile ``cushion'' where any transfer-based black-box attack can run on top of it (applied to that attack's chosen surrogate model) without change, yet reaping performance gains. The application of LRS-2 mirrors that of LRS-1.

The LRS approach is flexible whereby it allows the combined use of LRS-1 and LRS-2 as a ``double cushion.'' Achieving this simply involves a weighted sum of the two regularization terms applied to the loss function. We refer to this scenario as LRS-F. In our experiments, we demonstrate the enhanced performance of LRS-F.

\begin{table*}[ht!]
\begin{center}
\resizebox{0.8\linewidth}{!}{
\renewcommand{\arraystretch}{1}
\begin{tabular}{ccccccc}
\toprule
Method         & ResNet-50*              & VGG-19  & ResNet-152 & Inception v3 & DenseNet & MobileNet \\\midrule
PGD (2018)        &   {100.00\%}    & 39.22\% & 29.18\%    & 15.60\%      & 35.58\%  & 37.90\%   \\
TIM (2019)     &   {100.00\%} & 44.98\%          & 35.14\%          & 22.21\%          & 46.19\%          & 42.67\%          \\
SIM (2020)     &   {100.00\%} & 53.30\%          & 46.80\%          & 27.04\%          & 54.16\%          & 52.54\%          \\
LinBP (2020)   &   {100.00\%} & 72.00\% & 58.62\% & 29.98\% & 63.70\% & 64.08\%   \\
Admix (2021)  &   {100.00\%} & 57.95\%          & 45.82\%          & 23.59\%          & 52.00\%          & 55.36\%          \\
TAIG (2022)    &   {100.00\%}    & 54.32\% & 45.32\%    & 28.52\%      & 53.34\%  & 55.18\%   \\
ILA++ (2022)   &   99.96\%  & 74.94\% & 69.64\% & 41.56\% & 71.28\% & 71.84\%   \\\midrule

LRS-1 (ours)   & 100.00\% & 76.02\% & 72.36\% & 42.01\% & 71.23\% & {69.36\%} \\
LRS-2 (ours)   & 100.00\% & 78.24\% & 75.96\% & 46.14\% & 73.01\% & {73.45\%} \\
LRS-F (ours)   &  100.00\%  & \textbf{80.64\%} & \textbf{78.21\%} & \textbf{50.10\%} & \textbf{75.19\%} & {\textbf{76.24\%}}  \\
\bottomrule 
\\
\toprule
Method         & SENet   & ResNeXt & WRN     & PNASNet & MNASNet & \textbf{Average} \\\midrule
PGD (2018)         & 17.66\% & 26.18\% & 27.18\% & 12.80\% & 35.58\% & {27.69\%} \\
TIM (2019)     & 22.47\%           & 32.11\%          & 33.26\%          & 21.09\%          & 39.85\%          & {34.00\%}          \\
SIM (2020)     & 27.04\%           & 41.28\%          & 42.66\%          & 21.74\%          & 50.36\%          & {41.69\%}          \\
LinBP (2020)   & 41.02\% & 51.02\% & 54.16\% & 29.72\% & 62.18\% & {52.65\%} \\
Admix (2021)  & 30.28\%           & 41.94\%          & 42.78\%          & 21.91\%          & 52.32\%          & {42.40\%}          \\
TAIG (2022)    & 24.82\% & 38.36\% & 42.16\% & 17.20\% & 54.90\% & {41.41\%} \\
ILA++ (2022)   & 53.12\% & 65.92\% & 65.64\% & 44.56\% & 70.40\% & {62.89\%} \\\midrule

LRS-1 (ours)   & 54.27\% & 66.85\% & 67.21\% & 45.29\% & 72.03\% & {64.53\%} \\
LRS-2 (ours)   & 57.19\% & 69.48\% & 71.13\% & 48.39\% & 75.68\% & {67.57\%} \\
LRS-F (ours) &  \textbf{59.68\%}  & \textbf{71.96\%}& \textbf{74.61\%} & \textbf{52.43\%} & \textbf{76.87\%}  & {\textbf{69.91\%}}  \\
\bottomrule
\end{tabular}
} 
\caption{Attack success rates of SOTA transfer-based untargeted attacks on ImageNet using ResNet-50 as the surrogate model and PGD as the backend attack method, under the $\ell_\infty$ constraint with $\epsilon=8/255$. `*' denotes white-box attack.}
\label{tab:comapre_imagenet}
\end{center} 
\end{table*}

\section{Evaluation} \label{sec:experiments}

\subsection{Experiment Setup} 
{\bf Dataset.} We test untargeted $\ell_{\infty}$ black-box attacks on CIFAR-10 \cite{krizhevsky2009learning} and ImageNet \cite{russakovsky2015imagenet} datasets as the common benchmark \cite{dong2018boosting,dong2019evading,guo2020backpropagating,li2023making}. For CIFAR-10, we perform attacks on all test data. For ImageNet, we randomly sample 5,000 test images that are correctly classified by all the target models from the ImageNet validation set. Inputs to all models are re-scaled to $[0.0,1.0]$.

\textbf{Models under attack.} We take CIFAR-10 dataset for quick experiments verification, DenseNet \cite{Huang2017densely} is chosen as surrogate model due to its small model size and high classification accuracy, and five other networks serving as target (victim) models: VGG-19 with batch normalization \cite{Simonyan2015}, ResNet-18 \cite{He2016}, WRN \cite{Zagoruyko2016}, ResNeXt \cite{Xie2017aggregated}, PyramidNet  \cite{Han2017}. 
For ImageNet, we choose ResNet-50 \cite{He2016} as the surrogate model and 10 state-of-the-art classifiers as target victim models: VGG-19 \cite{Simonyan2015}, ResNet-152 \cite{He2016}, Inception v3 \cite{Szegedy2016}, DenseNet \cite{Huang2017densely}, MobileNet v2 \cite{Sandler2018mobilenetv2}, SENet \cite{Hu2018}, ResNeXt \cite{Xie2017aggregated}, WRN \cite{Zagoruyko2016}, PNASNet \cite{Liu2018}, and MNASNet \cite{Tan2019mnasnet}. For the above victim models, we follow their official pre-processing pipelines in our evaluation.

\textbf{Implementation details on ImageNet.} For LRS-1 regularization, we set $\lambda_1=5.0$, $h_1=0.01$. For LRS-2 regularization, we set $\lambda_2=5.0$, $h_2=1.5$. When use LRS-F as regularization, we keep the same $\lambda$ and $h$ values. We use an SGD optimizer with momentum 0.9 and weight decay 0.0005, the learning rate is fixed at 0.001, and the surrogate model is run for 10 epochs which is a tradeoff between efficiency and efficacy. With PGD as the back-end method, we run it for 50 iterations on ImageNet with perturbation range 8/255 and step size of 2/255. All experiments are performed on an NVIDIA V100 GPU.

\begin{table*}[ht!]
\begin{center}
\resizebox{0.8\linewidth}{!}{
\begin{tabular}{c|c|cccccc}
\toprule
Method          & DenseNet* & VGG19 & ResNet18 & WRN & ResNeXt   & PyramidNet   & {Average} \\\midrule
TIM (2019)      & {100.00\%} & 33.96\% & 23.46\% &56.49\% & 72.38\% & 23.14\%   & {41.89\%} \\
TIM+LRS-1      & {100.00\%} & 64.23\% & 53.19\% &81.03\% & 86.95\% & 50.62\%   & {67.80\%} \\
TIM+LRS-2      & {100.00\%} & 69.21\% & 57.39\% &86.98\% & 90.12\% & 55.13\%   & {71.17\%} \\
TIM+LRS-F      & {100.00\%} & \textbf{73.86\%} & \textbf{61.48\%} &\textbf{90.11\%} & \textbf{93.48\%} & \textbf{60.42\%}   & \textbf{75.87\%}\\\midrule

Admix (2021)   & {100.00\%} & 44.09\% & 34.80\% &64.36\% & 76.24\% & 27.65\%   & {49.43\%} \\
Admix+LRS-1   & {100.00\%} & 66.49\% & 58.96\% &85.69\% & 89.65\% & 55.48\%   & {71.05\%} \\
Admix+LRS-2   & {100.00\%} & 74.39\% & 63.59\% &88.94\% & 93.56\% & 62.47\%   & {76.39\%} \\
Admix+LRS-F   & {100.00\%} & \textbf{78.12\%} & \textbf{68.04\%} &\textbf{94.23\%} & \textbf{95.37\%} & \textbf{67.96\% }  & \textbf{80.14\%} \\\midrule

TAIG (2022)     & {100.00\%} & 41.69\% & 30.23\% &64.12\% & 75.89\% & 25.96\%    & {47.78\%} \\
TAIG+LRS-1     & {100.00\%} & 62.38\% & 51.29\% &80.33\% & 84.68\% & 51.46\%    & {66.03\%} \\
TAIG+LRS-2     & {100.00\%} & 73.18\% & 62.08\% &84.39\% & 92.04\% & 60.03\%    & {74.34\%} \\
TAIG+LRS-F     & {100.00\%} & \textbf{75.98\%} & \textbf{65.21\%} &\textbf{89.11\%} & \textbf{93.16\%} & \textbf{63.49\%}    & \textbf{77.99\%} \\
\bottomrule
\end{tabular}}
\caption{Attack success rates by combining SOTA transfer-based untargeted attacks with our methods, on CIFAR-10 using DenseNet as the surrogate model and PGD as the backbone attack method, under the $\ell_\infty$ constraint with $\epsilon=4/255$. `*' denotes white-box attack.}
\vspace{-5mm}
\label{tab:combine_cifar10}
\end{center} 
\end{table*}

\subsection{Experimental Results}
We conducted several sets of experiments in order to thoroughly evaluate the proposed approach. More experimental results are available in the supplementary material.

\textbf{Validation on small scale.} 
We first experiment on the relatively smaller CIFAR-10 using DenseNet as surrogate to evaluate LRS. 
In Table \ref{tab:quick_exp}, we compare adversarial transferability over the original pretrained surrogate model and that over LRS-transformed surrogate models, all using PGD as the base attack. The evaluation involved two perturbation scales $\epsilon$ \eqref{eq:1}. We observe that: (1) overall, applying LRS results in clear improvement by large margins; (2) LRS-2 boosts adversarial transferability more than LRS-1; (3) the best surrogate model is achieved by using both the first and second order regularization together, i.e., LRS-F, while at the cost of slightly higher computation overhead. Specifically, when $\epsilon=4/255$, we see an absolute value increase in the average attack success rate (ASR) of 20.03\%, 25.37\% and 30.10\% when the surrogate model is transformed by LRS-1, LRS-2 and LRS-F, respectively; when $\epsilon=8/255$, the corresponding improvements are 26.72\%, 28.86\% and 30.63\%, respectively. All of these are significant enhancements. In particular, when attacking PyramidNet with LRS-F transformed surrogate model under $\epsilon=8/255$, we achieved an increase of ASR by a remarkable 48.84\% in absolute value.

\textbf{Comparison with SOTA on large scale.}
We compare the attacking performance of LRS on 10 target models with state-of-the-art (SOTA) attacking methods, on the relatively large ImageNet dataset (the same comparison on CIFAR10 is reported in supplementary material). The SOTA attack methods for comparison include TIM \cite{dong2019evading}, SIM \cite{lin2020nesterov}, LinBP \cite{guo2020backpropagating}, Admix \cite{wang2021admix}, TAIG \cite{huang2022transferable} and ILA++ \cite{guo2022intermediate}. The results are presented in Table \ref{tab:comapre_imagenet}, which shows that all the LRS-cushioned attacking methods (LRS-1, LRS-2, LRS-F) outperform all the SOTA methods considerably. For example, looking at the Average ASR column of Table \ref{tab:comapre_imagenet}, LRS-F achieves an improvement over all the SOTA methods of between 7.02--35.91\%.

\textbf{Easily integrating with and supporting other attacks.}
As previously noted, LRS is a flexible ``cushion'' on which any other transfer-based black-box attack can execute without any change. In \tref{tab:combine_cifar10}, we report the results when applying LSR to TIM, Admix and ILA++ (besides PGD which has been shown). It can be seen that the transferability is enhanced significantly by 20--34\% on average due to the use of LRS. 

\textbf{Attacking ``secure'' models.}
For a more thorough evaluation, we also investigate how LRS performs when attacking DNN models that have been {\em adversarially trained} (and hence are much harder to attack). Once more, it showcases compelling performance. The detailed results are provided in supplementary material due to space limit.

\subsection{Exploring Further: Factors Enhancing Adversarial Transferability in Regularized Surrogate Models}

\textbf{Smaller local Lipschitz constant.}
A reduced Lipschitz constant indicates a smoother classifier. Therefore, we delve into whether our transformed surrogate models indeed exhibit increased smoothness through a smaller local Lipschitz constant. While computing the precise Lipschitz constant remains a open challenge, we can empirically gauge the surrogate models' local Lipschitzness using the empirical Lipschitz constant \cite{yang2020closer}:
\begin{align}\label{eq:lip_emp}
L_{emp} = \frac{1}{n} \sum_{i=1}^n \max _{\boldsymbol{x}_i^{\prime} \in \mathbb{B}_{\infty}\left(\boldsymbol{x}_i, \varepsilon\right)} \frac{\left\|f\left(\boldsymbol{x}_i\right)-f\left(\boldsymbol{x}_i^{\prime}\right)\right\|_2}{\left\|\boldsymbol{x}_i-\boldsymbol{x}_i^{\prime}\right\|_{2}}
\end{align}

We estimate this value using a PGD-like approach and calculate the average estimation across all test data points. Refer to \tref{tab:lip_const} for the empirical local Lipschitz constants. It clearly shows that our transformed surrogate models display significantly reduced local Lipschitz constants (by {\em more than an order of magnitude}). This contributes to a notably smoother loss landscape, minimizing the likelihood of the AE generation process being confined to undesirable local optima. Such optima yield low loss values yet possess complex non-smooth geometries that are challenging to navigate away from.

\begin{table}[ht!]
\small
\begin{center}
\begin{tabular}{|c|c|c|}
\hline
Surrogate model & DenseNet100 & ResNet50\\
\hline
Original pretrained  & 5.53 &  976.59\\
Transformed by LRS-1 & 0.79 &  57.62\\
Transformed LRS-2 & 0.67 &  53.21\\
Transformed LRS-F & \textbf{0.59} & \textbf{49.64} \\
\hline
\end{tabular}
\caption{Empirical local Lipschitz constant of surrogate model estimated via Eq.~\eqref{eq:lip_emp}. The constants of DenseNet and ResNet50 are evaluated on CIFAR10 and ImageNet, respectively.}
\label{tab:lip_const}
\vspace{-5mm}
\end{center}
\end{table}

\textbf{Smoother loss landscape.} 
Research has extensively explored flat optima's role in model generalization \cite{chaudhari2019entropy,keskar2017on,foret2020sharpness}, highlighting how optimizing weights toward flat optima can improve neural network generalization due to their robustness against shifts in the loss function between training and test data. In our context of developing first-order Lipschitz regularization, we propose that \emph{adversarial examples positioned within flat optima exhibit robustness against shifts in the loss function between surrogate and target models}, thus enhancing AE transferability.

To verify our hypothesis, we visualize the loss landscape of a surrogate model before and after transformation in \fref{fig:loss_landscape}. The original pretrained surrogate model features a highly non-linear and jagged loss surface. Conversely, regularization results in a notably smoother loss surface with flatter local optima. This visualization confirms our regularization strategy's effectiveness in smoothing out sharp optima in the loss landscape. Consequently, AEs generated using regularized surrogate models are more likely to reside within flat optima, boosting their transferability.



\textbf{More robust against attacks.}
Another perspective explaining the favorability of Lipschitz-regularized surrogate models for adversarial transferability is their increased robustness against adversarial attacks. When a neural network possesses a small Lipschitz constant, it signifies a strict control over changes in network output amidst input perturbations, leading to certified robustness guarantees \cite{finlay2018improved, zhang2022rethinking}. Consequently, generating AEs on such robust surrogate models enhances the effectiveness of the resulting AEs in deceiving less robust target models. This robustness contributes to adversarial transferability.

\begin{figure}[h!]
    \centering
    \includegraphics[width=\linewidth]{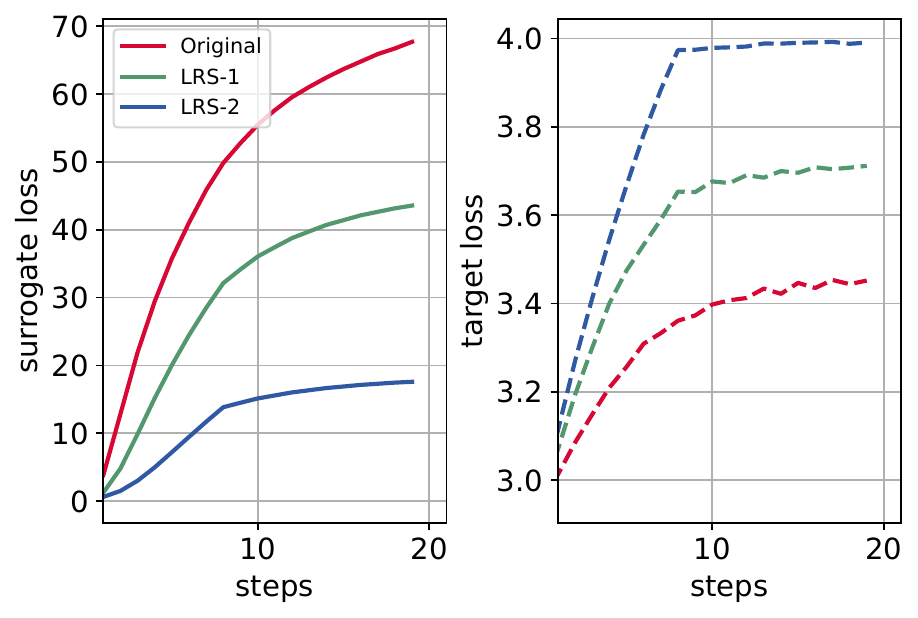}
    \caption{The loss of surrogate model (DenseNet) and target model (ResNet18), w.r.t. PGD-generated AE. It reveals that LRS-transformed models demonstrate more robustness and enable more transferable attacks.}
    \label{fig:loss}
\end{figure}

In line with this notion, \fref{fig:loss} illustrates that adversarial examples generated by PGD yield significantly lower losses on regularized surrogate models, indicating their enhanced robustness. However, their loss on target models is higher, signifying stronger black-box attacks and improved transferability.

\subsection{Ablation Studies}
We perform ablation studies on two crucial hyperparameters in our proposed LRS approach: the \emph{step size} and \emph{regularization coefficient}, denoted as $h_1$ and $\lambda_1$ for LRS-1, and $h_2$ and $\lambda_2$ for LRS-2. These parameters influence the locally enforced Lipschitz radius around the current AE. Larger values of $h_1$ and $h_2$ are generally preferred to enhance the neighborhood radius. Conversely, excessively large values can introduce finite difference method approximation errors, potentially misleading the AE update direction. The values of $\lambda_1$ and $\lambda_2$ serve to balance the trade-off between model accuracy and Lipschitz regularization.

\begin{figure}[h!]
    \centering
    \includegraphics[width=\linewidth]{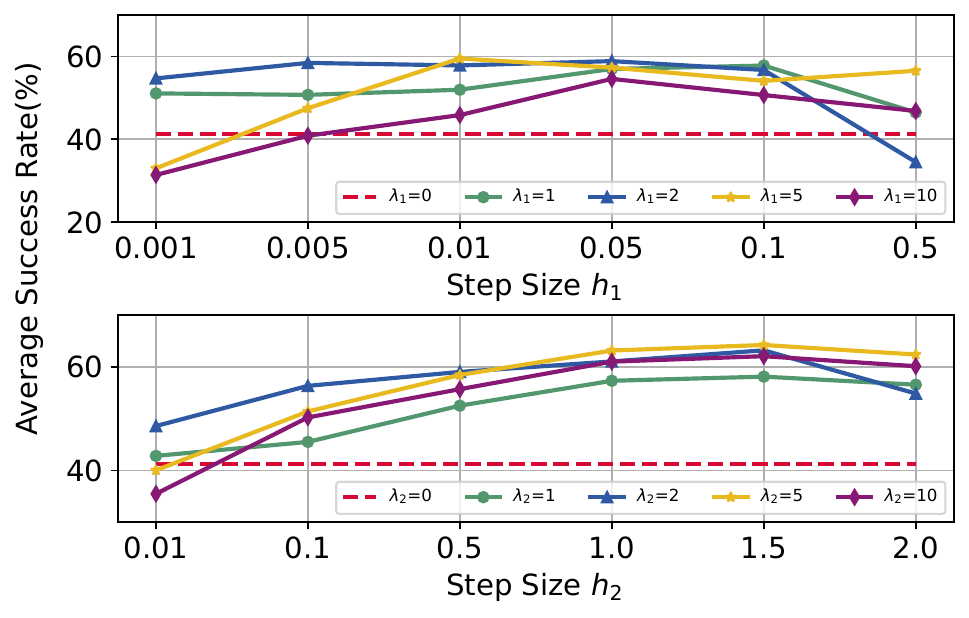}\vspace{-1mm}
    \caption{Ablation studies on average ASR under different hyperparameters $h$ and $\lambda$, the performance gains are consistent in a wide range of hyper-parameter values.}
    \label{fig:ablation}\vspace{-2mm}
\end{figure}

\fref{fig:ablation} presents the outcomes of our ablation studies. The ASR is computed by averaging over 5 target models on CIFAR10. The attacks are executed using PGD as the backend method with $\epsilon=4/255$. Our observations indicate that AE generated using LRS have significantly enhanced transferability compared to the case with $\lambda=0$. These performance improvements remain consistent across a reasonably broad range of $\lambda$ and $h$ values. This ablation study underscores the non-sensitive nature of LRS to hyperparameters, establishing its effectiveness across diverse conditions.

\section{Conclusion}  \label{sec:conclusion} 
This paper introduces a novel approach to enhancing adversarial transferability by transforming surrogate models via regularization, unlike in previous research where a pretrained model is chosen as is to serve as the (fixed) surrogate. We present Lipschitz Regularized Surrogate (LRS), a technique that imposes Lipschitz regularization to surrogate models for just a few training epochs. We show that this technique enables {\em any} existing transfer-based black-box AE generation method to produce highly transferable adversarial examples. This is validated through comprehensive experiments involving comparisons with numerous benchmark models, attack methods, and datasets. Our findings affirm the remarkable efficacy and superiority of LRS. Moreover, we offer insights into what and how properties of surrogate models promote adversarial transferability.

\section*{Acknowledgements}
This work was supported in part by the National Science Foundation (NSF) under Grant No. 2008878, and in part by the Air Force Research Laboratory (AFRL) and the Lifelong Learning Machines program by DARPA/MTO under Contract No. FA8650-18-C-7831. 
The research was also sponsored by the Army Research Laboratory and was accomplished under Cooperative Agreement Number W911NF-22-2-0209. 

\bibliography{aaai24}

\newpage

\begin{table*}[ht!]
\begin{center}
\resizebox{0.8\linewidth}{!}{
\begin{tabular}{c|c|cccccc}
\toprule
Method          & DenseNet* & VGG19 & ResNet18 & WRN & ResNeXt   & PyramidNet   & {Average} \\\midrule
PGD          & {100.00\%} & 29.79\% & 19.04\% & 54.41\% & 69.41\%   & 21.53\% & {38.84\%} \\
TIM (2019)      & {100.00\%} & 33.96\% & 23.46\% &56.49\% & 72.38\% & 23.14\%   & {41.89\%} \\
SIM (2020)      & {100.00\%} & 42.01\% & 32.06\% &59.26\% & 79.20\% & 26.46\%   & {47.60\%} \\
LinBP (2020)    & {100.00\%} & 53.26\% & 40.28\% &73.49\% & 83.14 \% & 38.96\%      & {57.83\%} \\
Admix (2021)   & {100.00\%} & 44.09\% & 34.80\% &64.36\% & 76.24\% & 27.65\%   & {49.43\%} \\
TAIG (2022)     & {100.00\%} & 41.69\% & 30.23\% &64.12\% & 75.89\% & 25.96\%    & {47.78\%} \\
ILA++ (2022) & {100.00\%} & 53.62\% & 41.46\% &73.19\% & 86.68\% & 36.49\%   & {58.49\%} \\\midrule
LRS-1 (Ours)  & 99.73\% & 55.97\% & 42.16\% & 72.66\% & 80.93\% & 42.64\% & 58.87\%\\
LRS-2 (Ours) & 99.82\% & 59.86\% & 48.98\% & 77.81\%  & 88.63\% & 46.78\% &  64.21\%\\
LRS-F  (Ours) & 99.93\% & \textbf{65.16\%} & \textbf{54.23\%} & \textbf{81.49\%}  & \textbf{92.76\%} & \textbf{51.07\%} & \textbf{68.94\%}\\
\bottomrule
\end{tabular}}
\caption{Attack success rates of SOTA transfer-based untargeted attacks on CIFAR-10 using DenseNet as the surrogate model and PGD as the backbone attack method, under the $\ell_\infty$ constraint with $\epsilon=4/255$. `*' denotes white-box attack.}
\label{tab:comapre_cifar10}
\end{center} 
\end{table*}

\section*{SUPPLEMENTARY MATERIALS}

\bigskip

\section{Comparison with SOTA on CIFAR10 dataset}

We compare the attacking performance of LRS on 5 target models with state-of-the-art (SOTA) attacking methods on CIFAR10 dataset. The SOTA attack methods for comparison include TIM \cite{dong2019evading}, SIM \cite{lin2020nesterov}, LinBP \cite{guo2020backpropagating}, Admix \cite{wang2021admix}, TAIG \cite{huang2022transferable} and ILA++ \cite{guo2022intermediate}. The results are presented in Table \ref{tab:comapre_cifar10}, which shows that all the LRS-cushioned attacking methods (LRS-1, LRS-2, LRS-F) outperform all the SOTA methods considerably. For example, looking at the Average ASR column of Table \ref{tab:comapre_cifar10}, LRS-F achieves an improvement over all the SOTA methods over 10.45\% by absolute value.

\section{Attacking ``Secure'' Models}

Besides attacking regularly trained models, we also conduct experiments on attacking adversarially trained models and models equipped with advanced defense methods---which are hence more ``secure''---for a thorougher evaluation of our proposed LRS method. We consider six such secure models, Inc-v $3_{\mathrm{ens} 3}$, Inc-v $3_{\mathrm{ens} 4}$ and IncRes-v $2_{\mathrm{ens}}$ \cite{tramer2018ensemble}, JPEG \cite{guo2018countering}, R\&P \cite{xie2018mitigating} and NRP \cite{naseer2020NRP}. Specifically, we craft adversarial examples on ResNet50 with $\epsilon=16/255$ on our selected ImageNet dataset and test the transferability on the six secure target models. The results are shown in Table \ref{tab:defense}. It shows that LRS remarkably improves the transferability of the backbone attack methods on all the six presumably more robust models. On average, the performance is improved by 23.80--27.48\% and 18.82--25.22\%, respectively, when using I-FGSM and TIM as the backbone attack methods.

\section{Comparison with More Methods}
It is also interesting to note that, we can sample checkpoints during the finetuning process and use those checkpoints to construct ensemble attacks like \cite{gubri2022lgv}, which will boost the black-box attack performance even further. Here we provide a comparison with LGV in Table \ref{tab:LGV}. LGV only works in emsemble setting since it achieves worse transferability if takes single model as surrogate. We sample multiple models during our finetuning as LGV and attack different checkpoints at every iteration. In the emsemble setting, our LRS-F outperform LGV by 6.7\% in terms of absolute value.

\begin{figure}[t]
\centering
\includegraphics[width=\linewidth]{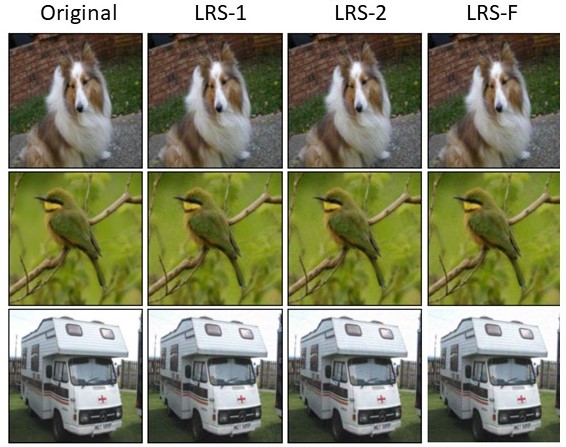}
\caption{Adversarial examples generated by our proposed LRS, showing that they are indistinguishable from original images by human eyes.}
\label{fig:Visualizations}
\end{figure}

\section{Visualizing Generated Adversarial Examples}

In Figure \ref{fig:Visualizations}, we show three randomly selected raw images and their corresponding adversarial examples crafted by our LRS methods. The adversarial examples are crafted on ResNet as source with PGD as the base attack and $\epsilon=8/255$. It shows that the adversarial noise added by our method are imperceptible to human eyes.

\section{Analysis of Computational Cost}
The LRS approach does not need any modification to the base methods that craft AE on the regularized surrogate models, as it acts as a flexible ``plug-in''. Compared to existing attack methods, the computational cost comes from the extra finetuning stage and is proportional to the total number of finetuning epochs. We have demonstrated that the finetuning is effective with as few as only 10 epochs. For example, on CIFAR10, the finetuning time of LRS-1, LRS-2 are about 2 and 5 minutes per epoch on one V100 GPU, respectively. Note that this finetuning process is performed {\em offline} and hence the overhead is affordable, as the most important performance indicator is the attacking success.

\begin{table*}[ht!] 

\begin{center}
\scalebox{0.95}{
\begin{tabular}{|c|c|cccccc|c|}
\hline
Source model & Attack & Inc-v3$_{ens3}$ & Inc-v3$_{ens4}$ & IncRes-v2$_{ens1}$ & JPEG & R\&P & NRP & Average\\
\hline\hline
\multirow{4}{*}{ResNet50} 
 & PGD &  4.3\% & 5.0\% & 2.3\% & 6.9\% & 2.2\% & 2.5\%&  3.87\%\\
 & PGD + LRS-1&  25.1\% & 24.7\% & 21.9\% & 38.6\% & 26.3\% & 29.4\% & 27.67\%\\
 & PGD + LRS-2&  26.7\% & 25.3\% & 23.0\% & 41.2\% & 28.4\% & 31.8\% & 29.40\%\\
 & PGD + LRS-F & \textbf{28.3\%}  & \textbf{27.6\%}  & \textbf{25.2\%}  &  \textbf{44.1\%} & \textbf{29.8\%} & \textbf{33.1\%} & \textbf{31.35\%}\\
 \hline
 \multirow{4}{*}{ResNet50} 
 & TIM &  21.3\% & 19.8\% & 12.6\% & 33.6\% & 18.4\% & 15.2\% & 20.15\%\\
 & TIM + LRS-1&  40.8\% & 41.0\% & 26.9\% & 51.4\% & 39.6\% & 34.1\% & 38.97\%\\
 & TIM + LRS-2&  43.6\% & 45.2\% & 30.4\% & 55.3\% & 43.5\% & 37.1\% & 42.52\%\\
 & TIM + LRS-F & \textbf{45.7\%}  & \textbf{48.2\%}  & \textbf{32.8\%}  &  \textbf{59.4\%} & \textbf{45.8\%} & \textbf{40.3\%} & \textbf{45.37\%}\\
\hline
\end{tabular}
}
\caption{Attacking ``secure'' models which underwent adversarial training or are equipped with advanced defense methods.}
\label{tab:defense}
\end{center}
\end{table*}

\begin{table*}[ht!]
\begin{center}
\resizebox{0.8\linewidth}{!}{
\begin{tabular}{|c|c|c|ccccc|c|}
\hline
$\epsilon$ & Methods & DenseNet* & VGG19 & ResNet18 & WRN & ResNeXt   & PyramidNet  & Average  \\
\hline\hline
 \multirow{5}{*}{4/255} & LGV & 99.98\% & 60.38\% & 48.96\% & 85.26\% & 87.64\% & 50.36\% & 66.52\% \\
 & LRS-1  & 99.85\% & 60.76\% & 47.89\% & 80.49\% & 85.69\% & 48.45\% & 64.66\%\\
 & LRS-2  & 99.92\% & 65.49\% & 53.27\% & 84.11\%  & 90.89\% & 52.06\% &  69.16\%\\
 & LRS-F  & 99.96\% & \textbf{69.86\%} & \textbf{58.98\%} & \textbf{87.15\%}  & \textbf{94.26\%} & \textbf{55.86\%} & \textbf{73.22\%}\\
\hline
\end{tabular}
}
\vspace{-0.2cm}
\caption{\footnotesize Compare LGV and LRS on CIFAR10.}
\label{tab:LGV}
\end{center}
\vspace{-1.0cm}
\end{table*}

\end{document}